# PRINCIPAL MANIFOLDS AND GRAPHS IN PRACTICE: FROM MOLECULAR BIOLOGY TO DYNAMICAL SYSTEMS

**ALEXANDER N. GORBAN**
*Department of Mathematics, University of Leicester, University Road*
*Leicester LE1 7RH, UK*
*E-mail: ag153@le.ac.uk*

**ANDREI ZINOVYEV**
*Institut Curie, U900 INSERM/Curie/Mines Paritech, rue d'Ulm 26*
*Paris, 75005, France*
*E-mail: andrei.zinovyev@curie.fr*

We present several applications of non-linear data modeling, using principal manifolds and principal graphs constructed using the metaphor of elasticity (elastic principal graph approach). These approaches are generalizations of the Kohonen's self-organizing maps, a class of artificial neural networks. On several examples we show advantages of using non-linear objects for data approximation in comparison to the linear ones. We propose four numerical criteria for comparing linear and non-linear mappings of datasets into the spaces of lower dimension. The examples are taken from comparative political science, from analysis of high-throughput data in molecular biology, from analysis of dynamical systems.

## 1. Introduction

In this paper, we provide several examples of application of the principal manifold and graphs methodology taken from applied projects: comparative political science, data analysis in molecular biology, theoretical methods of dynamical systems analysis. As it will be shown, one of the most common problems in these fields is how to approximate a finite set $D$ in $R^m$ for relatively large $m$ by a finite subset of a regular low-dimensional object in $R^m$.

The first hypothesis we have to check is: whether the dataset $D$ is situated near a low–dimensional affine manifold (plane) in $R^m$. If we look for a point, straight line, plane, ... that minimizes the average squared distance to the datapoints, we immediately come to the Principal Component Analysis (PCA).

PCA is one of the most seminal inventions in data analysis. Now it is textbook material. Nonlinear generalization of PCA is a great challenge, and many attempts have been made to answer it. In 1982 Kohonen introduced a type of *neural networks* called Self-Organizing Maps (SOM)[1].

With the SOM algorithm we take a finite metric space $Y$ with metric $\rho$ (a space of "neurons") and try to map it into $R^m$ with (a) the best preservation of initial structure in the image of $Y$ and (b) the best approximation of the dataset $D$. The SOM algorithm has several setup variables to regulate the compromise between these goals. We start from some initial approximation of the map, $\varphi_1{:}M \to R^m$. On each ($k$-th) step of the algorithm we have a datapoint $x \in D$ and a current approximation $\varphi_k{:}M \to R^m$. For these $x$ and $\varphi_k$ we define an "owner neuron" of $x$ in $Y$: $y_x = \mathrm{argmin}_{y\in Y} \|x-\varphi_k(y)\|$. The next approximation, $\varphi_{k+1}$, is

$$\varphi_{k+1}(y) = \varphi_k(y)+h_k\, w(\rho(y,y_x))(x-\varphi_k(y)). \qquad (1)$$

Here $h_k$ is a step size, $0 \le w(\rho(y,y_x)) \le 1$ is a monotonically decreasing cutting function and $\rho(y,y_x)$ is distance between $y$ and $y_x$. There are many ways to combine steps (1) in the whole algorithm. The idea of SOM is very flexible and seminal, has plenty of applications and generalizations, but, strictly speaking, we don't know what we are looking for: we have the algorithm, but no independent definition: SOM is a result of the algorithm work. The attempts to define SOM as solution of a minimization problem for some energy functional were not very successful[2].

For a known probability distribution, *principal manifolds* were introduced as lines or surfaces passing through "the middle" of the data distribution[3].

This intuitive vision was transformed into the mathematical notion of self-consistency: every point x of the principal manifold M is a conditional expectation of all points *z* that are projected into *x*. Neither manifold, nor projection should be linear: just a differentiable projection $\pi$ of the data space (usually it is $R^m$ or a domain in $R^m$) onto the manifold $M$ with the self-consistency requirement for conditional expectations: $x = \mathbf{E}(z|\pi(z) = x)$. For a finite dataset $D$, only one or zero datapoints are typically projected into a point of the principal manifold. In order to avoid overfitting, we have to introduce smoothers that become an essential part of the principal manifold construction algorithms.

SOMs give the most popular approximations for principal manifolds: we can take for $Y$ a fragment of a regular $k$-dimensional grid and consider the resulting SOM as the approximation to the $k$-dimensional principal manifold (see, for example, Refs. 4-5). Several original algorithms for construction of principal curves[6] and surfaces for finite datasets were developed during last decade, as well as many applications of this idea. In 1996, in a discussion about SOM at the 5th Russian National Seminar in Neuroinformatics, a method of multidimensional data approximation based on elastic energy minimization was proposed (see Refs. 7-14 and the bibliography there). This method is based on the analogy between the principal manifold and the elastic membrane (and plate). Following the metaphor of elasticity, we introduce two quadratic smoothness penalty terms. This allows one to apply standard minimization of quadratic functionals (i.e., solving a system of linear algebraic equations with a sparse matrix).

## 2. Method of principal elastic graphs and principal elastic maps

### 2.1. *Principal graphs and manifolds*

Here we give a short description of the structure of the elastic functional used for construction of principal graphs and manifolds. For more detailed description, see other works (in particular, Ref. 14).

In a series of works (see Refs. 7-16), the authors of this paper used metaphor of elastic membrane and plate to construct one-, two- and three-dimensional principal manifold approximations of various topologies. Mean squared distance approximation error combined with the elastic energy of the membrane serves as a functional to be optimised. The elastic map algorithm is extremely fast at the optimisation step due to the simplest form of the smoothness penalty. It is implemented in several programming languages as software libraries or front-end user graphical interfaces freely available from the web-site http://bioinfo.curie.fr/projects/vidaexpert. The software found applications in microarray data analysis, visualization of genetic texts, visualization of economical and sociological data and other fields[7-16].

Let $G$ be a simple undirected graph with set of vertices $V$ and set of edges $E$.

**Definition**. *k-star* in a graph $G$ is a subgraph with $k + 1$ vertices $v_{0,1,\dots,k} \in V$ and $k$ edges $\{(v_0, v_i)|i = 1, .., k\}$

**Definition**. Suppose that for each $k \geq 2$, a family $S_k$ of $k$-stars in $G$ has been selected. Then we define an *elastic graph* as a graph with selected families of $k$-stars $S_k$ and for which for all $E^{(i)} \in E$ and $S_k^{(j)} \in S_k$, the corresponding elasticity moduli $\lambda_i > 0$ and $\mu_{kj} > 0$ are defined.

**Definition**. *Primitive elastic graph* is an elastic graph in which every non-terminal node (with the number of neighbours more than one) is associated with a $k$-star formed by *all* neighbours of the node. All $k$-stars in the primitive elastic graph are selected, i.e. the $S_k$ sets are completely determined by the graph structure.

**Definition**. Let $E^{(i)}(0)$, $E^{(i)}(1)$ denote two vertices of the graph edge $E^{(i)}$ and $S_k^{(j)}(0),\dots, S_k^{(j)}(k)$ denote vertices of a $k$-star $S_k^{(j)}$ (where $S_k^{(j)}(0)$ is the central vertex, to which all other vertices are connected). Let us consider a map $\phi: V \to \mathbf{R}^m$ which describes an embedding of the graph into a multidimensional space. The *elastic energy of the graph embedding in the Euclidean space* is defined as

$$U^{\phi}(G) := U_E^{\phi}(G) + U_R^{\phi}(G), \qquad (2)$$

$$U_E^{\phi}(G) := \sum_{E^{(i)}} \lambda_i \left\| \phi(E^{(i)}(0)) - \phi(E^{(i)}(1)) \right\|^2, \quad (3)$$

$$U_E^{\phi}(G) := \sum_{S_k^{(j)}} \mu_{kj} \parallel \phi(S_k^{(j)}(0)) - \frac{1}{k}\sum_{i=1}^{k} \phi(S_k^{(j)}(i)) \parallel^2. \quad (4)$$

**Definition**. *Elastic map* is a continuous manifold $Y \in \mathbf{R}^m$ constructed from the elastic net as its grid approximation using some between-node interpolation procedure. This interpolation procedure constructs a continuous mapping $\phi_c:\{\phi_1,\dots,\phi_{\dim(G)}\} \to \mathbf{R}^m$ from the discrete map $\phi: V \to \mathbf{R}^m$, used to embed the graph in $\mathbf{R}^m$, and the discrete values of node indices $\{\lambda_1^i,\dots,\lambda_{\dim(G)}^i\}$, $i = 1\dots|V|$. For example, the simplest *piecewise linear elastic map* is build by piecewise linear map $\phi_c$.

**Definition**. *Elastic principal manifold of dimension s* for a dataset $X$ is an elastic map, constructed from an elastic net $Y$ of dimension $s$ embedded in $\mathbf{R}^m$ using such a map $\phi_{opt}: Y \to \mathbf{R}^m$, that corresponds to the minimal value of the functional

$$U^{\phi}(X,Y) = \mathrm{MSD}_W(X,Y) + U^{\phi}(G), \quad (5)$$

where the weighted mean squared distance from the dataset $X$ to the elastic net $Y$ is calculated as the distance to the finite set of vertices $\{\mathbf{y}^1=\phi(v_1),..., \mathbf{y}^k=\phi(v_k)\}$.

In the Euclidean space one can apply an EM algorithm for estimating the elastic principal manifold for a finite dataset. It is based in turn on the general algorithm for estimating the locally optimal embedding map $\phi$ for an arbitrary elastic graph $G$ (see Ref. 14).

### 2.2. *Pluriharmonic graphs as ideal approximators*

Approximating datasets by one dimensional principal curves is not satisfactory in the case of datasets that can be intuitively characterized as *branched*. A principal object which naturally passes through the 'middle' of such a data distribution should also have branching points that are missing in the simple structure of principal curves. Introducing such branching points converts principal curves into *principal graphs*.

In Refs. 10,12,14 it was proposed to use a universal form of non-linearity penalty for the branching points. The form of this penalty is defined in the previous section (4) for the elastic energy of graph embedment. It naturally generalizes the simplest three-point *second derivative* approximation squared: for a 2-star (or rib) the penalty equals $\| \phi(S_2^{(j)}(0)) - \frac{1}{2}(\phi(S_2^{(j)}(1)) + \phi(S_2^{(j)}(2))) \|^2$, for a 3-star it is $\| \phi(S_3^{(j)}(0)) - \frac{1}{3}(\phi(S_3^{(j)}(1)) + \phi(S_3^{(j)}(2)) + \phi(S_3^{(j)}(3))) \|^2$, etc.

For a $k$-star this penalty equals to zero iff the position of the central node coincides with the mean point of its neighbors. An embedment $\phi(G)$ is 'ideal' if all such penalties equal to zero. For a *primitive elastic graph* this means that this embedment is a *harmonic function on graph*: its value in each non-terminal vertex is a mean of the value in the closest neighbors of this vertex.

For non-primitive graphs we can consider stars which include not all neighbors of their centers. For example, for a square lattice we create elastic graph (elastic net) using 2-stars (ribs): all vertical 2-stars and all horizontal 2-stars. For such elastic net, each non-boundary vertex belongs to two stars. For a general elastic graph $G$ with sets of $k$-stars $S_k$ we introduce the following notion of pluriharmonic function.

**Definition.** A map $\phi: V \to \mathbf{R}^m$ defined on vertices of $G$ is *pluriharmonic* iff for any $k$-star $S_k^{(j)} \in S_k$ with the central vertex $S_k^{(j)}(0)$ and the neighbouring vertices $S_k^{(j)}(i)$, $i = 1...k$, the equality holds:

$$\phi(S_k^{(j)}(0)) = \frac{1}{k}\sum_{i=1}^{k} \phi(S_k^{(j)}(i)). \quad (6)$$

Pluriharmonic maps generalize the notion of linear map and of harmonic map, simultaneously. For example:

1) 1D harmonic functions are linear;
2) If we consider an $n$-dimensional cubic lattice as a primitive graph (with $2n$-stars for all non-boundary vertices), then the correspondent pluriharmonic functions are just harmonic ones;
3) If we create from $n$-dimensional cubic lattice the standard $n$-dimensional elastic net with 2-stars (each non-boundary vertex is a center of $n$ 2-stars, one 2-star for each coordinate direction), then pluriharmonic functions are linear.

In the theory of principal curves and manifolds the penalty functions were introduced to penalise deviation from linear manifolds (straight lines or planes). We proposed to use pluriharmonic embeddings as "ideal objects" instead of manifolds and to introduce penalty (5) for deviation from this ideal form.

### 2.3. *Complexity of principal graphs*

The principal graphs can be called *data approximators of controllable complexity*. By complexity of the principal objects we mean the following three notions:

1) *Geometric complexity*: how far a principal object deviates from its ideal configuration; for the elastic principal graphs we explicitly measure deviation from the 'ideal' pluriharmonic graph by the elastic energy $U_\phi(G)$ (3) (this sort of complexity is often just a *measure of non-linearity*);
2) *Structural complexity measure*: it is some non-decreasing function of the number of vertices, edges and $k$-stars of different orders $\mathrm{SC}(G)=\mathrm{SC}(|V|,|E|,|S_2|,\ldots,|S_m|)$; this function penalises for number of structural elements;

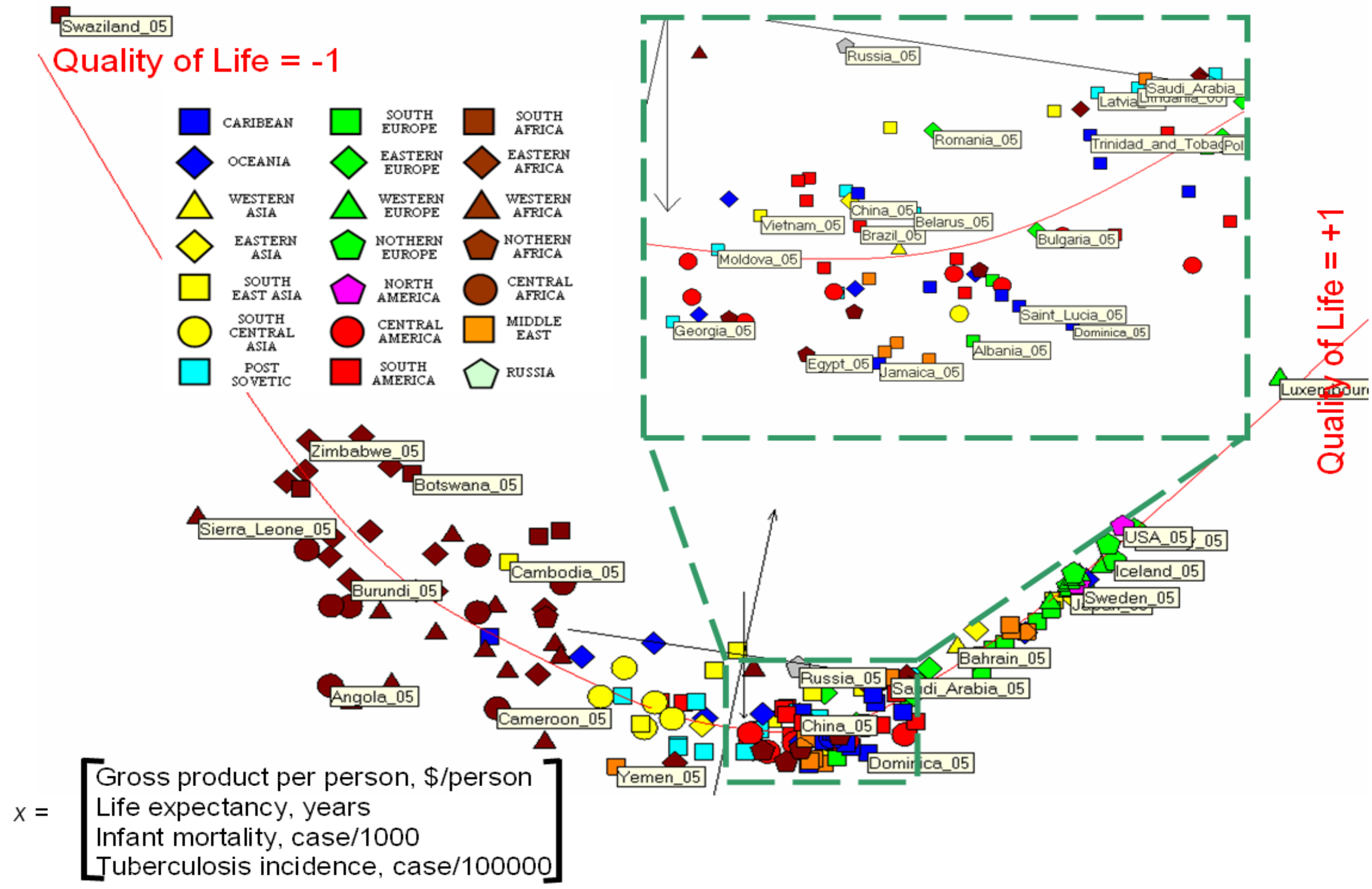

Fig. 1. Approximation of data used for constructing the quality of life index. Each of 192 points represents a country in 4-dimensional space formed by the values of 4 indicators (gross product, life expectancy, infant mortality, tuberculosis incidence). Different forms and colors correspond to various geographical locations. Red line represents the principal curve, approximating the dataset. Three arrows show the basis of principal components. The best linear index (first principal component) explains 76% of variation, while the non-linear index (principal curve) explains 93% of variation.

3) *Construction complexity* is defined with respect to a graph grammar as a minimum number of applications of elementary transformations necessary to construct given $G$ from the simplest graph (one vertex, zero edges).

The construction complexity is defined with respect to a *grammar* of elementary transformation. The graph grammars[17,18] provide a well-developed formalism for the description of elementary transformations. An elastic graph grammar is presented as a set of production (or substitution) rules. Each rule has a form $A \to B$, where $A$ and $B$ are elastic graphs. When this rule is applied to an elastic graph, a copy of $A$ is removed from the graph together with all its incident edges and is replaced with a copy of $B$ with edges that connect $B$ to the graph. For a full description of this language we need the notion of a *labeled graph*. Labels are necessary to provide the proper connection between $B$ and the graph[18]. An approach based on *graph grammars* to constructing effective approximations of *elastic principal graph* was proposed recently[10,12,14].

Let us define *graph grammar* $O$ as a set of graph grammar operations $O=\{o_1,..,o_s\}$. All possible applications of a graph grammar operation $o_i$ to a graph $G$ gives a set of transformations of the initial graph $o_i(G) = \{G_1, G_2, \ldots, G_p\}$, where $p$ is the number of all possible applications of $o_i$ to $G$. Let us also define a sequence of $r$ different graph grammars $\{O^{(1)} = \{o_1^{(1)},\ldots,o_{s_1}^{(1)}\}, \cdots, O^{(r)} = \{o_1^{(r)},\ldots,o_{s_r}^{(r)}\}\}$.

Let us choose a grammar of elementary transformations, predefined boundaries of structural complexity $SC_{\max}$ and construction complexity $CC_{\max}$, and elasticity coefficients $\lambda_i$ and $\mu_{kj}$.

**Definition**. *Elastic principal graph* for a dataset $X$ is such an elastic graph $G$ embedded in the Euclidean space by the map $\phi: V \to \mathbf{R}^m$ that $\mathrm{SC}(G) \le SC_{\max}$, $\mathrm{CC}(G) \le CC_{\max}$, and $U_\phi(G) \to \min$ over all possible elastic graphs $G$ embeddings in $\mathbf{R}^m$.

For the simplest choice of grammar, this definition gives us *principal trees* (see the section 3.3).

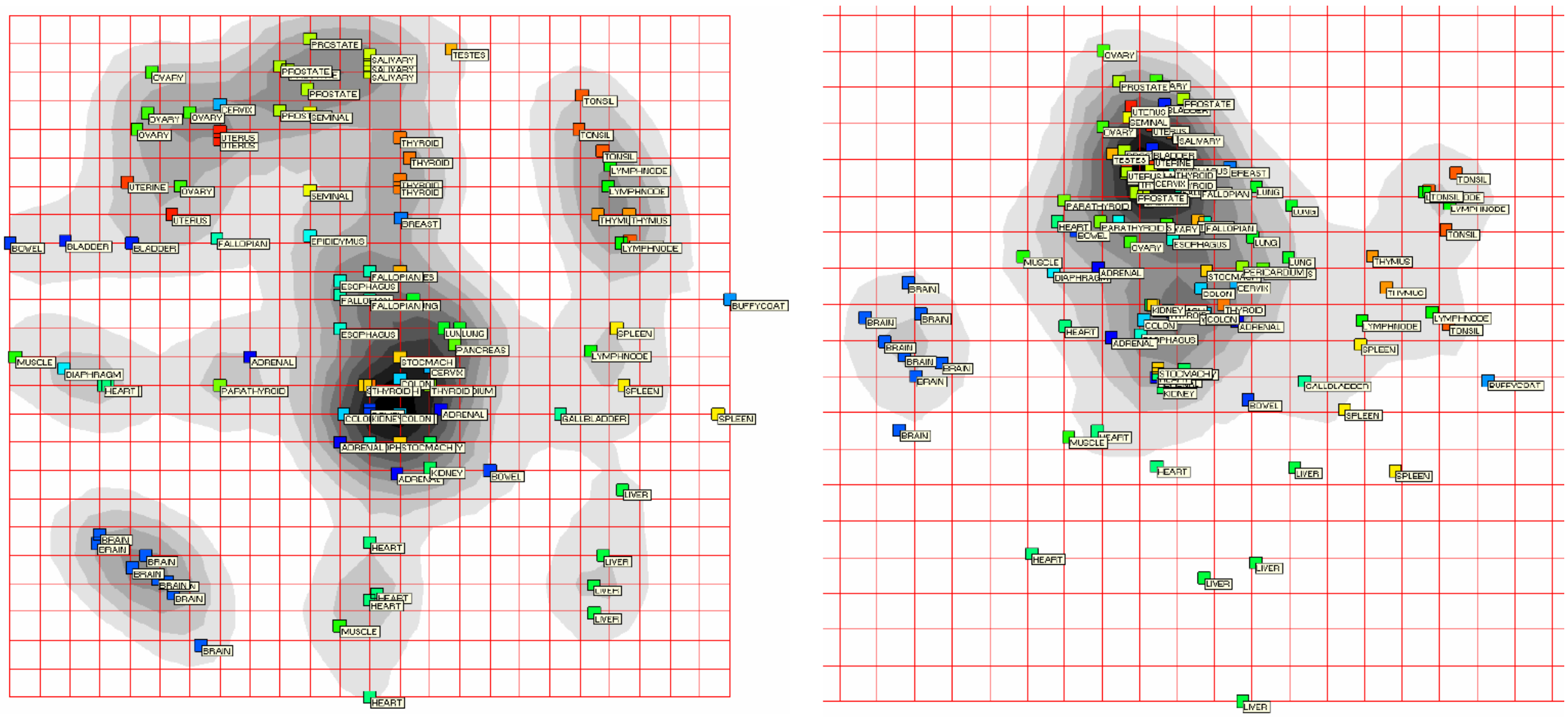

a) ELMAP2D b) PCA2D

Fig. 2. Comparing linear (b) and non-linear (a) two-dimensional projections of data on the expression of genes in normal human tissues. Each point represents a tissue sample (brain, heart, pancreas, etc., each denoted by a distinct color) by a vector of expression of several thousands of genes. The linear projection is done by the standard PCA approach. The non-linear projection is made by the elastic map method, described in this paper.

## 3. Real-life examples of principal graph and manifold applications

### 3.1. *Happiness (quality of life) is non-linear*

Let us take a simple real-life example where application of linear data approximation is not satisfactory and non-linear one-dimensional principal manifolds (principal curves) are needed to adequately describe the data.

In the Political World Atlas project, launched by the Moscow State Institute of International Relations[19], systematic quantitative data on 192 modern countries for the 1989-2005 period were collected. One of the goals was to estimate the "objective" position of post-sovetic countries, in particular, in terms of "objective" estimates of the quality of life. Similarly to other projects on quantification of country happiness or life quality, the following quantitative indicators could be used for constructing an integral index and ranking: gross product per capita in dollars, life expectancy in years, infant mortality in cases per 10000 habitants, tuberculosis incidence in cases per 100000 habitants. A common practice in this field is to combine these or similar features in one index (number) which usually represents a linear combination of features with weights determined by experts (see, for example, Ref. 20). This approach is inevitably biased by the experts' opinion. An unbiased approach would be to use the best one-dimensional approximation of data to introduce the "objective" ranking. It makes sense if a good approximation function exists.

Each of 192 countries is characterized by the 4 values, thus, we can represent this dataset as a cloud of 192 points in 4-dimensional space. Simple visualization of this distribution allows to make a conclusion that *no linear function exists that can equally well serve for reducing the dimension of this space from 4 to 1 (see Fig.1).* The distribution is intrinsically curved, hence, any linear mapping will inevitably give strong distortions in one or other region of dataspace.

There is a simple reason for this. Observing Fig.1, it is easy to realize that all countries can be roughly separated in two groups. First group consists of very wealthy countries, mostly from Western and Northern Europe, USA, Australia and some others (right branch of the distribution on the Fig.1). It happens that the most of variation among these countries can be attributed to the gross product per capita feature while others are approximately equal for them and do not contribute significantly to the variance. The second group (left branch of the distribution on the Fig.1) consists of very poor countries (mostly, African), which are “equally poor” in terms of the gross product per capita but can be very different in terms of their problems (lower of higher life expectancy, level of infectious diseases, conditions of the state health system), for a number of reasons (wars, difference in internal politics, etc.). However, this classification is not perfect: in fact, there are many countries which are localized in the non-linear junction between these two branches of the 4-dimensional distribution. This intermediate group includes the most of the post-sovetic countries which are the subject of the politologic analysis in this study, hence, it is important to correctly position them in the global picture.

To do this, we have constructed a principal curve approximating the curved data distribution (red line on the Fig. 1). The advantage of using the principal curve instead of principal line was significant in term of Mean-Squared Error (93% vs 76% of explained variance). Importantly, using the non-linear index for non-linear one-dimensional dimension reduction can significantly change the absolute and relative rankings for many countries, including Russia. Thus, for the index, constructed using the first principal component, Russia in 2005 is ranked the 86$^{th}$ (with the most prosperous Luxembourg at the 1$^{st}$ place), after, for example, Iran (74$^{th}$) and Egypt (84$^{th}$). For non-linear index, Russia is ranked the 71$^{st}$ before Iran (77$^{th}$) and Egypt (85$^{th}$).

### 3.2. *Dimension reduction for microarrays*

Now let us consider application of two-dimensional principal manifolds for data visualization. This is their natural application since they permit to create a mapping from multidimensional space of data to two- or three- dimensional spaces, thus, providing a possibility to create visual images of multidimensional distributions.

DNA microarray data is a rich source of information for molecular biology (for a recent overview, see Ref. 21). This technology found numerous applications in understanding various biological processes including cancer. It allows simultaneous screening of the expression of all genes in a cell exposed to some specific conditions (for example, stress, cancer, treatment, normal conditions). Obtaining a sufficient number of observations (chips), one can construct a table of "samples vs genes", containing logarithms of the expression levels of, typically several thousands ($n$) of genes, in typically several tens ($m$) of samples.

For this study, we took three distinct microarray datasets, provided at http://www.math.le.ac.uk/people/ag153/homepage/PrincManLeicAug2006.htm:

1) Breast cancer dataset from Ref. 22, 286 tumor samples and 17816 genes, several natural groupings of samples accordingly to survival of patients, status of some cell receptors and molecular classification of cells.

2) Bladder cancer dataset from Ref. 23, 40 tumor samples, several natural groupings of samples accordingly to the stage and grade of the tumor

3) Collection of microarrays for normal human tissues from Ref. 24, 103 samples of healthy tissues, grouped accordingly to the tissue of origin (brain, heart, pancreas, etc.).

In Fig. 2 we compare data visualization scatters after projection of the normal tissues dataset, provided in Ref. 24, onto the *linear* two-dimensional and *non-linear* two-dimensional principal manifold. The later one is constructed by the elastic maps approach. Each point here represents a sample of tissue, labeled by the tissue name. Before dimension reduction it is represented as a vector in $\mathbf{R}^n$, containing the expression values for all $n$ genes in the sample.

One of the conclusions that can be made from the Fig. 2 is that the non-linear mapping is capable of resolving the structure of the data distribution in more details. In particular, some groups of points that are not separated on the linear scatter become separated on the non-linear one. We have decided to quantify this effect, thus, comparing the quality of *linear and non-linear mappings of the same dimension.* Four criteria, each characterizing a projection of dataset into the low-dimensional space were developed:

1) *Mean-Squared Error (MSE)*. This is simply the mean squared distance from all of the points to the manifold, where the projection is orthogonal (by the shortest distance). MSE is measured in % of total variance. Notice that non-linear approximations should give smaller MSE by construction, since they are less rigid than the linear ones.

2) *Quality of distance mapping (QDM)*. This is a correlation coefficient between the pair-wise distances between data points before ($d_{ij}$) and after ($\hat{d}_{ij}$) projection onto the manifold:

$$QDM = corr(d_{ij}, \hat{d}_{ij}) \qquad (7)$$

For estimating *QDM* measure we also proposed to calculate the correlation coefficient only on "the most representative" subset of pair-wise distances, selected by a procedure which we have called "Natural PCA". The procedure was introduced in Ref. 13. The first "natural" component is defined by the pair of the most distant points ($i1, j1$). The second component is such a pair of points ($i2, j2$) that 1) $i2$ is the most distant point to the first component, where the distance from a point to a set of points $S$ is defined as the distance to the closest point in $S$; 2) $j2$ is the point from the first component which is the closest to $i2$. All the next components are introduced analogously: the $n$th component ($i_n$, $j_n$) is such a pair of points that the $i_n$ is the most distant from the union $S_{n-1}$ of all components from 1 to $n$-1 and $j_n$ is the point from this union $S_{n-1}$ which is the closest to $i_n$. We use both Pearson and Spearman correlation coefficients for calculating QDM.

3) *Quality of point neighborhood preservation (QNP)*. For measuring it, for every data point $i$ we calculate the size of the intersection of the set of $k$ neighbours in the multi-dimensional space $S(i;k)$ and in the low-dimensional space $\hat{S}(i;k)$. QNP is the average value of this intersection divided by $k$:

$$QNP_k = 1/k \sum_{i=1..N} |S(i;k) \cap \hat{S}(i;k)| / N \; . \qquad (8)$$

4) *Quality of group compactness (QGC)*. In this test we assume that there is a label C($i$) associated with every point $i$. Then, for each label $B$, we can calculate the average number of points of the same color in the $k$-neighborhood of the points before and after projection. Let us define $c(i;k)$ as the number of points in the $k$-neighbourhood of the point $i$ having the label $C(i)$. Then, for a label $B$,

$$QGC_k(B) = 1/k \sum_{C(i)=B} c(i;k) / N(B) \, , \qquad (9)$$

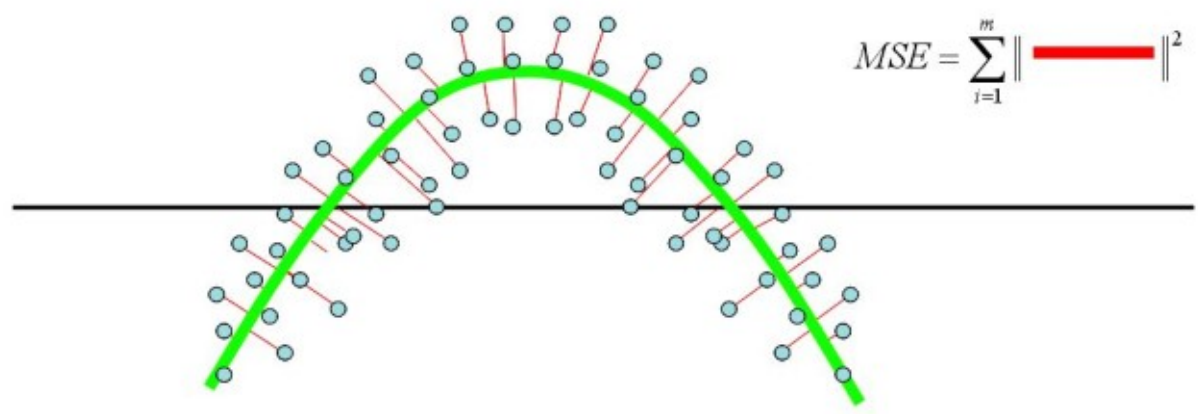

| Dataset | ELMAP2D | PC1 | PC2 | PC3 | PC4 | PC5 | PC10 |
|---|---|---|---|---|---|---|---|
| Breast cancer MSE | 48.1 | 52.6 | 50.7 | 49.3 | 48.4 | 47.6 | 45.3 |
| Variation explained | - | 7.9% | 14.3% | 19.0% | 22.0% | 24.6% | 31.5% |
| Bladder cancer MSE | 40.6 | 48.7 | 45.4 | 42.8 | 40.7 | 39.2 | 33.0 |
| Variation explained | - | 21.0% | 31.4% | 38.9% | 44.8% | 48.8% | 63.8% |
| Normal tissues MSE | 36.9 | 48.8 | 45.5 | 42.3 | 40.1 | 38.5 | 32.4 |
| Variation explained | - | 10.7% | 19.1% | 26.0% | 30.3% | 32.2% | 40.7% |

Fig. 3. Criterion 1 (**Mean-Squared Error, MSE**) for comparison of linear (of dimension 1-5, 10, columns PC1-PC5, PC10) and non-linear projections of data (of dimension 2, column ELMAP2D). The color mark on the table shows the comparative performance of the non-linear method compared to the linear ones. Here two-dimensional ELMAP approach performs as well as 4-dimensional linear approximations.

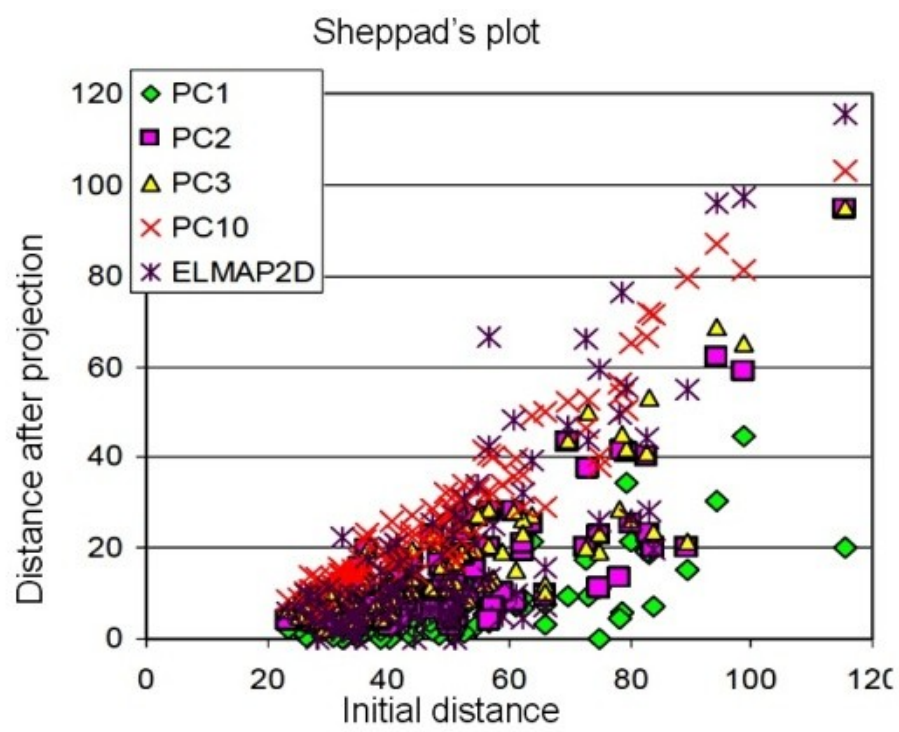

| Dataset/method | ELMAP2D | PC1 | PC2 | PC3 | PC4 | PC5 | PC10 |
|---|---|---|---|---|---|---|---|
| Breast cancer/Pearson | 0.60 | 0.40 | 0.52 | 0.61 | 0.65 | 0.69 | 0.75 |
| Breast cancer/Spearman | 0.40 | 0.19 | 0.32 | 0.36 | 0.42 | 0.49 | 0.56 |
| Bladder cancer/Pearson | 0.87 | 0.82 | 0.84 | 0.88 | 0.89 | 0.91 | 0.96 |
| Bladder cancer/Spearman | 0.68 | 0.57 | 0.60 | 0.70 | 0.70 | 0.75 | 0.90 |
| Normal tissues/Pearson | 0.80 | 0.68 | 0.78 | 0.82 | 0.86 | 0.87 | 0.95 |
| Normal tissues/Spearman | 0.68 | 0.56 | 0.69 | 0.79 | 0.84 | 0.86 | 0.94 |

Fig. 4. Criterion 2 (**Quality of distance mapping, QDM**). For the column titles see the legend to the Fig. 3. The plot shows the comparison of pair-wise point distances in the initial mutli-dimensional space and on the manifold after projection. The subset of distances selected by NPCA approach is shown (see the text for explanations).

where $N(B)$ is the number of points having the label $B$. Values of QGC close to one would indicate that the points of the label $B$ forms a very compact and separated from the other colors group. Comparing QGC(B) before and after projection into the low-dimensional space, one can conclude on what happened with the group of points of the label $B$.

It is clear from the Fig. 3-6 that the non-linear two-dimensional principal manifolds provide systematically better results accordingly to all four criteria, achieving the performance of three- and four- dimensional linear principal manifolds.

A special attention should be made to the performance of the non-linear principal manifolds with respect to the QGC criterion. It works particularly well for the collection of normal tissues. There are cases when neither linear nor non-linear low-dimensional manifolds could put together points of the same class and there are a few examples when linear manifolds perform better. In the latter cases (Breast cancer's A, B, lumA, lumB and "unclassified" classes, bladder cancer T1 class), almost all class compactness values are close sto the estimated random values which means that these classes have big intra-class dispersions or are poorly separated from the others. In this case the value of class compactness becomes unstable (look, for example, at the classes A and B of the breast cancer dataset) and depends on random factors which can not be taken into account in this framework.

The closer class compactness is to unity, the easier one can construct a decision function separating this class from the others. However, in the high-dimensional space, due to many degrees of freedom, the "class compactness" might be compromised and become better after appropriate dimension reduction. In Fig. 6 one can find examples when dimension reduction gives better class compactness in comparison with that calculated in the initial multi-dimensional space (breast cancer basal subtype, bladder cancer Grade 2 and T1, T2+ classes). It means that sample classifiers can be regularized by dimension reduction using PCA-like methods.

There are several particular cases (breast cancer basal subtype, bladder cancer T2+, Grade 2 subtypes) when non-linear manifolds give better class compactness than both the multidimensional space and linear principal manifolds of the same dimension. In these cases we can conclude that the dataset in the regions of these classes is naturally "curved" and the application of non-linear techniques for classification regularization is an appropriate solution.

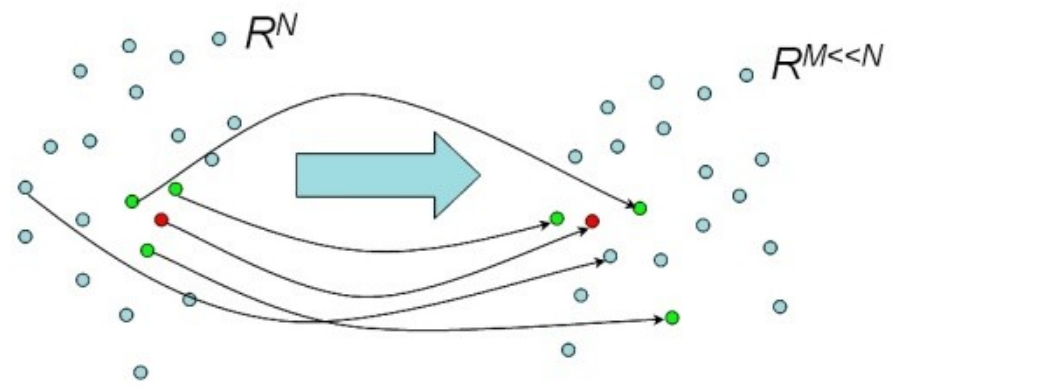

| Dataset | ELMAP2D | PC1 | PC2 | PC3 | PC4 | PC5 | PC10 | RANDOM |
|---|---|---|---|---|---|---|---|---|
| Breast cancer (k=10) | 0.26 | 0.13 | 0.20 | 0.28 | 0.31 | 0.38 | 0.47 | 0.04 ± 0.06 |
| Bladder cancer (k=5) | 0.53 | 0.34 | 0.53 | 0.61 | 0.64 | 0.70 | 0.80 | 0.12 ± 0.14 |
| Normal tissues (k=5) | 0.49 | 0.23 | 0.33 | 0.43 | 0.50 | 0.54 | 0.69 | 0.05 ± 0.09 |

Fig. 5. Criterion 3 (**Quality of point neighborhood preservation**, **QNP**). For the column titles see the legend to the Fig. 3. The RANDOM column shows the random permutation test when random $k$ points are taken instead of the real $k$-neighbourhood.

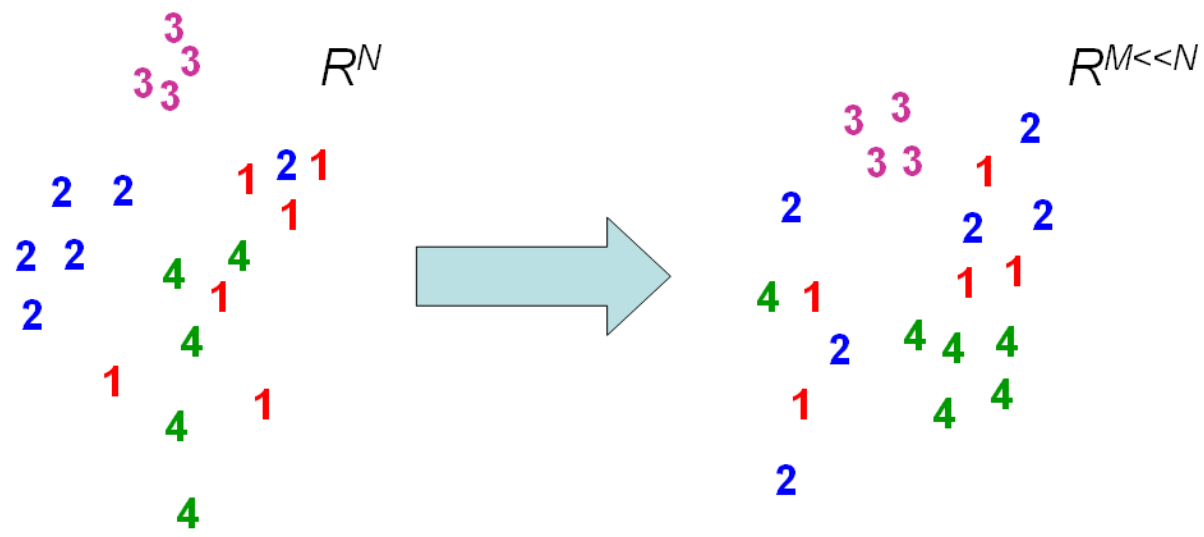

| | Class | ND | ElMap2D | PC1 | PC2 | PC3 | PC4 | PC5 | PC10 | Random | |
|---|---|---|---|---|---|---|---|---|---|---|---|
| | Breast cancer (GROUP), k=5 | | | | | | | | | | |
| | A (193 samples) | 0.73 | 0.67 | 0.71 | 0.67 | 0.68 | 0.69 | 0.70 | 0.70 | 0.67±0.27 | 1 |
| | B (93 samples) | 0.31 | 0.30 | 0.33 | 0.29 | 0.35 | 0.37 | 0.38 | 0.31 | 0.33±0.21 | 1 |
| | Breast cancer (TYPE), k=3 | | | | | | | | | | |
| - | lumA (95 samples) | 0.61 | 0.60 | 0.64 | 0.65 | 0.67 | 0.72 | 0.71 | 0.74 | 0.33±0.27 | 1 |
| ++ | basal (55 samples) | 0.88 | 0.93 | 0.83 | 0.86 | 0.88 | 0.90 | 0.88 | 0.89 | 0.19±0.17 | 3 |
| - | unclas. (42 samples) | 0.28 | 0.20 | 0.27 | 0.25 | 0.27 | 0.27 | 0.30 | 0.29 | 0.15±0.16 | 1 |
| + | normal (35 samples) | 0.68 | 0.28 | 0.14 | 0.19 | 0.31 | 0.41 | 0.41 | 0.55 | 0.12±0.19 | 2 |
| ++ | errb2 (34 samples) | 0.62 | 0.36 | 0.24 | 0.34 | 0.32 | 0.32 | 0.43 | 0.59 | 0.12±0.19 | 2 |
| - | lumB (25 samples) | 0.21 | 0.08 | 0.20 | 0.20 | 0.21 | 0.25 | 0.23 | 0.36 | 0.09±0.17 | 1 |
| | Bladder cancer (TYPE), k=3 | | | | | | | | | | |
| | Ta (20 samples) | 0.85 | 0.83 | 0.80 | 0.85 | 0.85 | 0.85 | 0.85 | 0.85 | 0.50±0.29 | 3 |
| | T1 (11 samples) | 0.58 | 0.67 | 0.45 | 0.69 | 0.67 | 0.70 | 0.67 | 0.63 | 0.27±0.26 | 4 |
| ++ | T2+ (9 samples) | 0.85 | 0.89 | 0.11 | 0.85 | 0.81 | 0.89 | 0.78 | 0.81 | 0.22±0.24 | 4 |
| | Bladder cancer (GRADE), k=3 | | | | | | | | | | |
| ++ | Grade 3 (32 samples) | 0.94 | 0.92 | 0.86 | 0.83 | 0.90 | 0.95 | 0.94 | 0.94 | 0.8±0.23 | 3 |
| + | Grade 2 (6 samples) | 0.5 | 0.61 | 0.22 | 0.39 | 0.67 | 0.67 | 0.67 | 0.61 | 0.15±0.23 | 4 |
| | Normal tissues (TISSUE), k=2 | | | | | | | | | | |
| | BRAIN (8) | 1.0 | 1.0 | 1.0 | 0.94 | 1.0 | 1.0 | 1.0 | 1.0 | 0.08±0.19 | 3 |
| +++ | HEART (6) | 0.92 | 0.50 | 0.17 | 0.25 | 0.25 | 0.25 | 0.33 | 0.58 | 0.06±0.16 | 2 |
| +++ | THYROID (6) | 0.92 | 0.67 | 0.17 | 0.17 | 0.25 | 0.17 | 0.67 | 0.75 | 0.06±0.17 | 3 |
| +++ | PROSTATE (5) | 1.0 | 0.8 | 0.0 | 0.0 | 0.1 | 0.4 | 0.5 | 1.0 | 0.05±0.16 | 3 |
| +++ | OVARY (5) | 0.7 | 0.6 | 0.2 | 0.1 | 0.3 | 0.4 | 0.5 | 0.7 | 0.05±0.15 | 3 |
| - | LUNG (4) | 0.88 | 0.40 | 0.5 | 0.63 | 0.75 | 0.63 | 0.63 | 0.75 | 0.04±0.13 | 2 |
| | ADRENAL (4) | 0.625 | 0.0 | 0.0 | 0.0 | 0.0 | 0.25 | 0.5 | 0.625 | 0.04±0.14 | 2 |
| ++ | LIVER (4) | 0.88 | 0.88 | 0.13 | 0.75 | 0.75 | 0.88 | 0.88 | 0.88 | 0.04±0.14 | 3 |
| +++ | SALIVARY (4) | 1.0 | 1.0 | 0.25 | 0.25 | 0.50 | 0.88 | 1.0 | 1.0 | 0.04±0.14 | 3 |
| ++ | TONSIL (4) | 0.88 | 0.63 | 0.0 | 0.13 | 0.34 | 0.75 | 0.75 | 0.88 | 0.04±0.14 | 3 |
| | LYMPHNODE (4) | 0.75 | 0.25 | 0.25 | 0.25 | 0.0 | 0.25 | 0.25 | 0.25 | 0.04±0.14 | 2 |
| +++ | FALLOPIAN (4) | 0.88 | 0.50 | 0.0 | 0.13 | 0.25 | 0.25 | 0.25 | 0.75 | 0.04±0.14 | 3 |
| + | SPLEEN (3) | 0.67 | 0.50 | 0.50 | 0.33 | 0.33 | 0.89 | 0.67 | 0.83 | 0.03±0.12 | 3 |
| + | KIDNEY (3) | 1.0 | 0.33 | 0.0 | 0.0 | 0.33 | 0.33 | 0.33 | 0.67 | 0.03±0.12 | 2 |
| | UTERUS (3) | 0.83 | 0.33 | 0.0 | 0.33 | 0.33 | 0.33 | 0.33 | 0.83 | 0.03±0.12 | 2 |
| | BOWEL (3) | 0.17 | 0.0 | 0.0 | 0.0 | 0.0 | 0.0 | 0.0 | 0.0 | 0.03±0.12 | 1 |
| | ESOPHAGUS (3) | 0.67 | 0.0 | 0.0 | 0.0 | 0.0 | 0.0 | 0.0 | 0.67 | 0.03±0.12 | 2 |
| | COLON (3) | 0.33 | 0.0 | 0.17 | 0.0 | 0.0 | 0.17 | 0.17 | 0.17 | 0.03±0.12 | 1 |
| | STOCMACH (3) | 0.5 | 0.0 | 0.0 | 0.0 | 0.0 | 0.0 | 0.0 | 0.0 | 0.03±0.12 | 2 |

Fig.6. Criterion 4 (**Quality of group compactness, QGC**). For the column titles see the legend to the Fig. 3. The numbers on the right and the row colors have the following meaning (see the graph above): number 3 is a group which is compact in the original space and remains compact after the projection; number 1 group is not compact before and after projection; number 2 group being compact before projection, becomes dispersed after the projection; number 4 group being not compact before, becomes compact after the projection. '+' on the left means 'better than linear', '++' means 'much better', '+++' means 'incomparably better', '-' means 'worse'.

We can conclude that non-linear principal manifolds provide systematically better or equal resolution of class separation in comparison with linear manifolds of the same dimension. They perform particularly well when there are many small and relatively compact heterogeneous classes (as in the case of normal tissue collection).

### 3.3. *Principal trees and Metro Maps*

Let us demonstrate how the idea of graph grammars (section 2) allows constructing the simplest non-trivial type of the principal graphs, called *principal trees*[12,14]. For this purpose let us introduce a simple 'Add a node, bisect an edge' graph grammar (see Fig. 7) applied to the class of primitive elastic graphs.

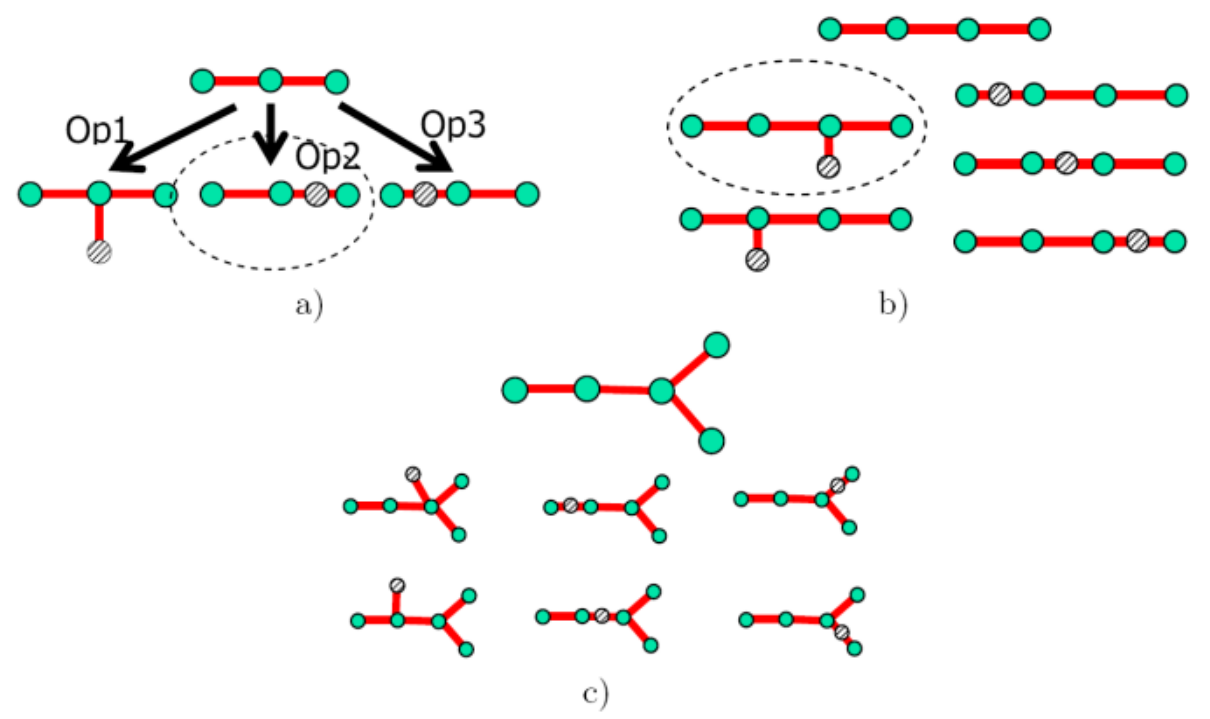

Fig. 7. Illustration of the simple "add node to a node" or "bisect an edge" graph grammar. a) We start with a simple 2-star from which one can generate three distinct graphs shown. The "Op1" operation is adding a node to a node, operations "Op1" and "Op2" are edge bisections (here they are topologically equivalent to adding a node to a terminal node of the initial 2-star). For illustration let us suppose that the "Op2" operation gives the biggest elastic energy decrement, thus it is the "optimal" operation. b) From the graph obtained one can generate 5 distinct graphs and choose the optimal one. c) The process is continued until a definite number of nodes are inserted.

**Definition**. *Principal tree* is an acyclic primitive elastic principal graph.

**Definition**. *'Add a node, bisect an edge' graph grammar* $O^{(grow)}$ applicable for the class of primitive elastic graphs consists of two operations: 1) The transformation *"add a node"* can be applied to any vertex $v$ of $G$: add a new node $z$ and a new edge $(v, z)$; 2) The transformation *"bisect an edge"* is applicable to any pair of graph vertices $v$, $v'$ connected by an edge $(v, v')$: delete edge $(v, v')$, add a vertex $z$ and two edges, $(v, z)$ and $(z, v')$. The transformation of the elastic structure (change in the star list) is induced by the change of topology, because the elastic graph is primitive. Consecutive application of the operations from this grammar generates trees, i.e. graphs without cycles.

**Definition**. *'Remove a leaf, remove an edge' graph grammar* $O^{(shrink)}$ applicable for the class of primitive elastic graphs consists of two operations: 1) The transformation '*remove a leaf*' can be applied to any vertex $v$ of $G$ with connectivity degree equal to 1: remove $v$ and remove the edge $(v,v')$ connecting v to the tree; 2) The transformation '*remove an edge*' is applicable to any pair of graph vertices $v$, $v'$ connected by an edge $(v, v')$: delete edge $(v, v')$, delete vertex $v'$, merge the $k$-stars for which $v$ and $v'$ are the central nodes and make a new $k$-star for which $v$ is the central node with a set of neighbors which is the union of the neighbors from the $k$-stars of $v$ and $v'$.

Also we should define the structural complexity measure $SC(G)=SC(|V|,|E|,|S_2|,\ldots,|S_m|)$. Its concrete form depends on the application field. Here are some simple examples:

1) $SC(G) = |V|$ : i.e., the graph is considered more complex if it has more vertices;
2) $$SC(G) = \begin{cases} |S_3|, \text{ if } |S_3| \le b_{\max} \text{ and } \sum_{k=4}^{m} |S_k| = 0 \\ \infty, \text{ otherwise} \end{cases},$$
i.e., only $b_{\max}$ simple branches (3-stars) are allowed in the principal tree.

To construct the principal tree, the following simple algorithm is applied:

1) Initialize the elastic graph $G$ by 2 vertices $v_1$ and $v_2$ connected by an edge. The initial map $\phi$ is chosen in such a way that $\phi(v_1)$ and $\phi(v_2)$ belong to the first principal line in such a way that all the data points are projected onto the principal line segment defined by $\phi(v_1)$, $\phi(v_2)$;

2) For a sequence of grammars $O^{(j)} = \{O^{(grow)}, O^{(grow)}, O^{(shrink)}\}$, $j$=1..3, repeat steps 3-6:

3) Apply all grammar operations from $O^{(j)}$ to $G$ in all possible ways; this gives a collection of candidate graph transformations $\{G_1, G_2, \ldots\}$;

4) Separate $\{G_1, G_2, \ldots\}$ into *permissible* and *forbidden* transformations; permissible transformation $G_k$ is such that $SC(G_k) \le SC_{\max}$ , where $SC_{\max}$ is some predefined structural complexity ceiling;

5) Optimize the embedding $\phi$ and calculate the elastic energy $U_\phi(G)$ of graph embedment for every permissible candidate transformation, and choose such a graph $G_{opt}$ that gives the minimal value of the elastic functional: $G_{opt} = \arg \inf_{G_k \in permissible\ set} U_\phi(G_k)$ ;

6) Substitute $G \leftarrow G_{opt}$ ;

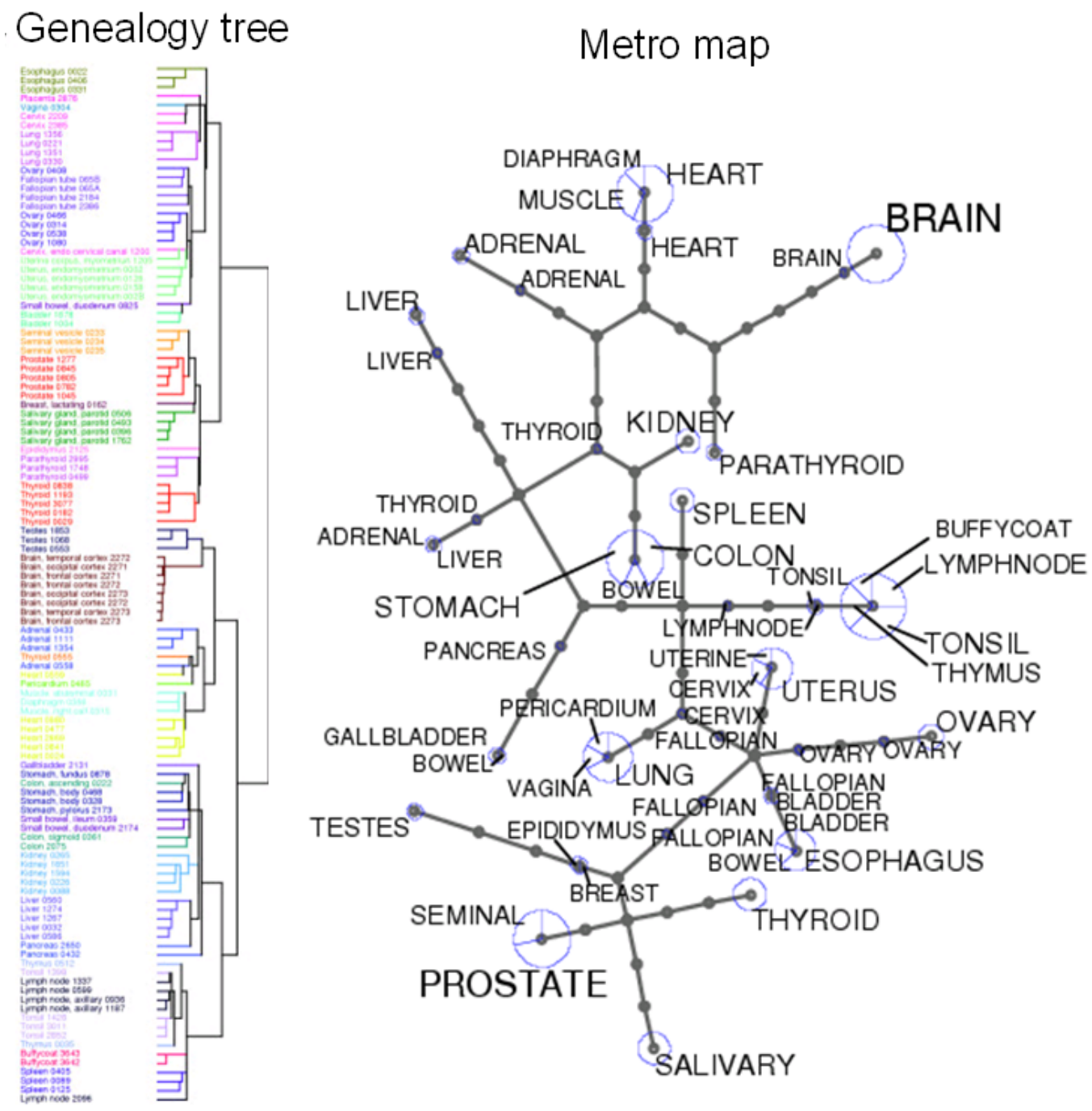

Fig. 8. “Metro map” representation of normal human tissue samples (on the right) and the hierarchical dendrogram of the same data (on the left, shown only for comparison). Both methods approximate the data by a tree-like structure, but using different metaphors (“genealogy tree” in the case of hierarchical dendrogram and “metro map” for the harmonic dendrite used by the principal tree method). In both representations the user can estimate a distance from one sample to another by summing up distances along the path in the tree. The size of circles on the metro map diagram is proportional to the number of points projected into the corresponding node, the pie diagram represents the composition of the cluster in terms of pre-existing sample groups (human tissues, in this case).

7) Repeat steps 2-6 until the set of permissible transformations is empty or the number of operations exceeds a predefined number – the construction complexity.

Using the ‘tree trimming’ grammar $O^{(shrink)}$ allows to produce principal trees closer to the global optimum, trimming excessive tree branching and fusing $k$-stars separated by small ‘bridges’.

Principal trees can have applications in data visualization. A principal tree is embedded into a multidimensional data space. It approximates the data so that one can project points from the multidimensional space into the closest node of the tree. The tree by its construction is a one-dimensional object, so sthis projection performs dimension reduction of the multidimensional data. The question is how to produce a planar tree layout? Of course, there are many ways to layout a tree on a plane without edge intersection. But it would be useful if both local tree properties and global distance relations would be represented using the layout. We can require that

1) In a two-dimensional layout, all $k$-stars should be represented equiangular; this is the small penalty configuration;

2) The edge lengths should be proportional to their length in the multidimensional embedding; thus one can represent between-node distances.

This defines a tree layout up to global rotation and scaling and also up to changing the order of leaves in every $k$-star. We can change this order to eliminate edge intersections, but the result cannot be guaranteed. In order to represent the global distance structure, it was found that a good approximation for the order of $k$-star leaves can be taken from the projection of every $k$-star on the linear principal plane[12] calculated for all data point or on the local principal plane in the vicinity of the $k$-star, calculated only for the points close to this star. The resulting layout can be further optimized using some greedy optimization methods.

Note that the distance on the metro map is estimated by summing up the lengths of branches along the path (Fig.8). Hence, any layout of the tree will not distort this information. The $k$-stars are projected onto the principal plane in order to find a good ordering of nodes inside the star, as a heuristics, to avoid excessive tree branch intersections. However, this ordering does not encode the distance along the tree branches but rather provide a way to generate a nice 2D layout.

The point projections are then represented as pie diagrams, where the size of the diagram reflects the number of points projected into the corresponding tree node. The sectors of the diagram allow us to show proportions of points of different classes projected into the node. An example of the metro map representation applied to the case of microarray dataset of normal tissues is shown in Fig. 8 (compare with 2D representation of the same dataset shown in Fig. 2).

### 3.4. *Principal manifolds for dynamical systems analysis*

*Invariant manifold* is a central concept in the theory of dynamical systems and its construction provides a consistent approach for model reduction[25]. The general picture behind this concept is that starting from an initial condition, the dynamical system quickly reaches vicinity of a curved low-dimensional manifold and during most of its dynamics remains close to it.

In Ref. 26 we constructed a dynamical model of NFkB biochemical signaling cascade. The model contains 35 dynamic variables representing concentrations of various biochemical species. This system is capable for sustained oscillations and converges to a limit cycle in the phase space. When studying the details of the dynamics of this system, we found that the dynamics in the vicinity of the limit cycle can be characterized by presence of a two-dimensional invariant manifold[27].

Using the same methodology as described above in the section 2 and a phase space sampling technique, we have constructed the invariant manifold approximation. To do it, we first sample the trajectories of the dynamical system in the vicinity of its limit cycle in the following way:

1) Using the PCA approach we found a linear manifold in which the cycle trajectory is embedded.

2) A system trajectory was computed in some interval of time $[0;t_m]$, started from a randomly chosen point of the limit cycle plus random shift in the linear manifold calculated at Step 1. The size of this shift $\delta$ can be made up to the border of the phase space (when one of the concentrations becomes zero).

3) The trajectory obtained at Step 2 was cut into two parts: for $t\in[0;\alpha t_m]$ and $t\in[\alpha t_m;t_m]$ intervals, where $\alpha\in[0;1]$. For the sampling, only the second part of the trajectory was used, properly discretized. It was done to skip the first (fast) dynamics of the system towards a hypothetical invariant manifold.

4) Steps 2 and 3 were repeated until sufficient (50000 in our experiments) number points were sampled.

The resulting distribution of points each representing a ‘snapshot’ along the trajectory of the dynamical system was approximated by a non-linear two-dimensional principal manifold constructed using the elastic map approach. The resulting manifold projected into the subspace of the first three (linear) principal components is shown on the Fig. 9. The constructed approximation to the invariant manifold can be further used for model reduction by projection of the system dynamics onto it[25,28].

Principal component analysis applied to the trajectories of a dynamical system sometimes called Karhunen-Loéve expansion[29,30] and it is a useful tool for reducing complexity of the dynamics. Hence, the approach presented in this example can be called the non-linear Karhunen-Loéve expansion.

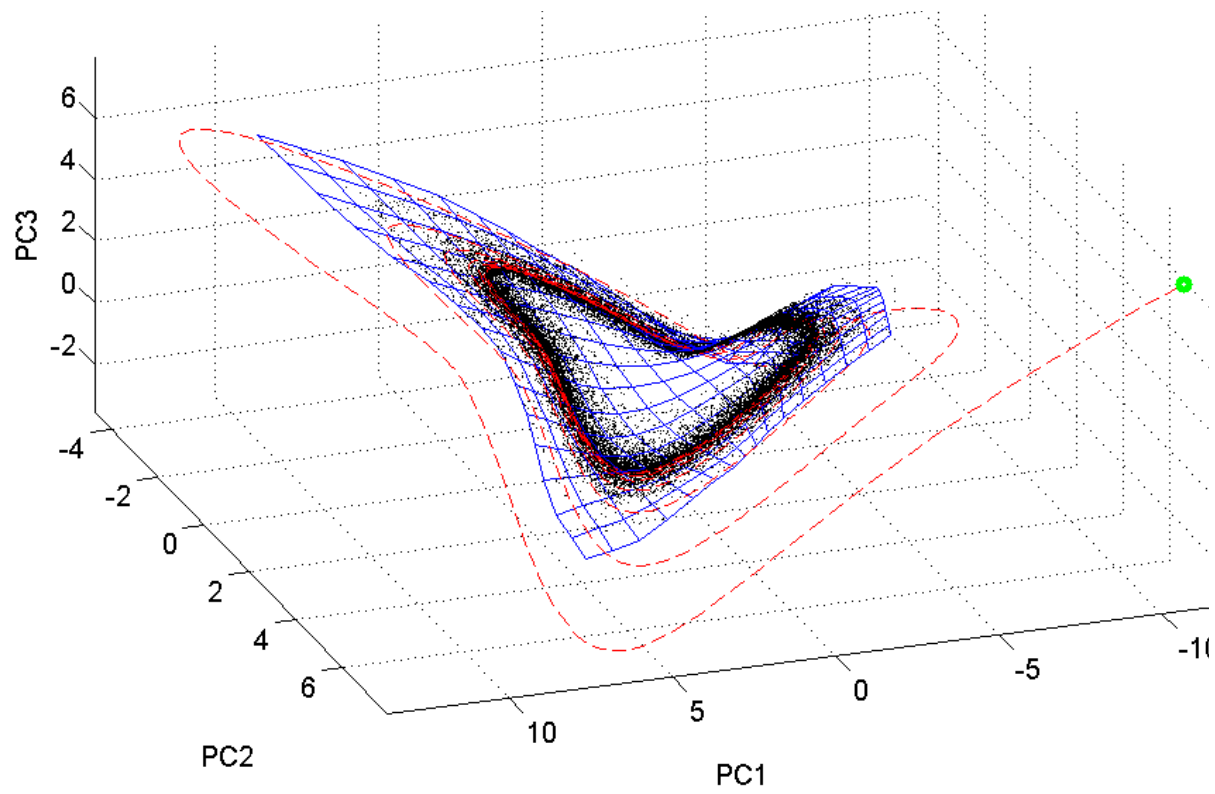

Fig. 9. Two-dimensional invariant manifold (blue grid) approximation obtained by application of the principal manifold methodology to the set of “snapshots” (black points) along trajectories of a dynamical system. The manifold is shown in the 3D projection of the first three principal components. By red one particular trajectory is shown, quickly approaching the manifold and further converging along it to the limit cycle.

## 4. Conclusions

Principal component analysis was introduced into applied science by Karl Pearson more than one hundred years ago[31] and since then it became one of the most used mathematical tool in many domains of science (in every domain, where approximation of a finite set of points is required).

Non-linear extensions of this method (Self-Organizing neural networks and their further generalizations such as principal manifolds, principal graphs and principal trees) also can serve as a universal tool allowing to approximate complex distributions of data points, when the linear approximation happens to be insufficient. To prove superiority and advantage of applying the non-linear approximations, one can use the set of benchmarking criteria described in this paper.

The elastic graph approach can be interpreted as an intermediate between absolutely flexible *neural gas*[32] and significantly more restrictive *Self-Organizing Maps* and *elastic maps*.

Using efficient implementation of principal graphs and manifolds provided by the elastic graph approach,

applying these methods in practice become relatively easy and computationally fast exercise. In this paper, we demonstrate this on several practical examples: from comparative political science, data analysis in molecular biology and analysis of dynamical systems for biochemical modeling.